\documentclass[review]{elsarticle}

\usepackage{lineno,hyperref}
\modulolinenumbers[5]

\journal{Information Fusion}
\usepackage{amssymb}
\usepackage{amsmath}
\usepackage{graphicx, subfigure}
\usepackage{url}
\usepackage{array}
\usepackage{verbatim}
\usepackage{slashbox}
\usepackage{algorithmic}
\usepackage{algorithm}
\usepackage{multirow}
\usepackage{booktabs}

\newtheorem{subsec:coding}{subsec:coding}

\begin{document}

\begin{frontmatter}
\title{AI-Skin : Skin Disease Recognition based on Self-learning and Wide Data Collection through a Closed Loop Framework}

\author[firstauthoraddr]{Min Chen\corref{cor1}}
\ead{minchen2012@hust.edu.cn}

\author[firstauthoraddr]{Ping Zhou}
\ead{pingzhou.cs@qq.com}

\author[secondauthoraddr,thirdauthoraddr]{Di Wu\corref{cor1}}
\ead{wudi27@sysu.edu.cn}

\author[firstauthoraddr]{Long Hu\corref{cor1}}
\ead{hulong@hust.edu.cn}

\author[fourthauthoraddr]{Mohammad Mehedi Hassan}
\ead{mmhassan@ksu.edu.sa}

\author[fifthauthoraddr]{Atif Alamri}
\ead{atif@ksu.edu.sa}

\cortext[cor1]{Corresponding authors: Min Chen, Di Wu, Long Hu}
\address[firstauthoraddr]{School of Computer Science and Technology,\\ Huazhong University of Science and Technology, Wuhan, China}
\address[secondauthoraddr]{School of Data and Computer Science, Sun Yat-sen University, Guangzhou, China}
\address[thirdauthoraddr]{Guangdong Province Key Laboratory of Big Data Analysis and Processing, Guangzhou, China}
\address[fourthauthoraddr]{Chair of Pervasive and Mobile Computing and Information Systems Department, \\ College of Computer and Information Sciences,
King Saud University, Riyadh 11543, Saudi Arabia}
\address[fifthauthoraddr]{Chair of Pervasive and Mobile Computing and Software Engineering Department, \\ CCIS, King Saud University, Riyadh 11543, Saudi Arabia}


\begin{abstract}
There are a lot of hidden dangers in the change of human skin conditions, such as the sunburn caused by long-time exposure to ultraviolet radiation, which not only has aesthetic impact causing psychological depression and lack of self-confidence, but also may even be life-threatening due to skin canceration. Current skin disease researches adopt the auto-classification system for improving the accuracy rate of skin disease classification. However, the excessive dependence on the image sample database is unable to provide individualized diagnosis service for different population groups. To overcome this problem, a medical AI framework based on data width evolution and self-learning is put forward in this paper to provide skin disease medical service meeting the requirement of real time, extendibility and individualization. First, the wide collection of data in the close-loop information flow of user and remote medical data center is discussed. Next, a data set filter algorithm based on information entropy is given, to lighten the load of edge node and meanwhile improve the learning ability of remote cloud analysis model. In addition, the framework provides an external algorithm load module, which can be compatible with the application requirements according to the model selected. Three kinds of deep learning model, i.e. LeNet-5, AlexNet and VGG16, are loaded and compared, which have verified the universality of the algorithm load module. The experiment platform for the proposed real-time, individualized and extensible skin disease recognition system is built. And the system's computation and communication delay under the interaction scenario between tester and remote data center are analyzed. It is demonstrated that the system we put forward is reliable and effective.
\end{abstract}


\end{frontmatter}


\section{Introduction}

A skin disease is the pathological state affecting the body surface. Long-time exposure to ultraviolet radiation or the radiation from high-frequency wireless equipment may induce the skin canceration. According to the statistical data report from American Cancer Society~\cite{ACSociety2018}, it is estimated that 91,270 new melanoma cases are diagnosed in the United States in 2018, and meanwhile it is estimated that about 9,320 people will die from melanoma. Melanoma has a high cure rate at early detection, with 99\% of 5-year relative survival rate. However, since it is easier to spread to other parts of the body than non-melanoma skin cancers, the 5-year relative survival rate at long-term stage drops to 20\%.

The symptom of skin diseases is a long and constantly changing process. Generally, the health care provider should provide assessment where changes have occurred in certain area of the skin for over a month or longer. However, due to multiple factors such as poor medical conditions and cumbersome medical process, patients always ignore such changes of their skin, or wrongly identify them as other skin injuries. Meanwhile, no family physician or supervision organization carries out the regular maintenance and treatment since it is not required to report medical records to the cancer registry.

The development of body sensor network~\cite{FortinoCollabo2015,CarolBSN2019}, artificial intelligence, cloud computing~\cite{XuHybrWor2018} and wireless network communication~\cite{JiangWiFiSen2018,HuPerioChar2019} has brought opportunities to the cognitive medical service~\cite{FridayFusi2019,TawalbehMo2016,GravinaAaaS2017}. The remote health monitoring, health guidance and feedback can be realized through multi-sensor data fusion~\cite{GravinaMfusion2017,KanjoFusion2018} with the help of remote medical devices, such as mobile phone~\cite{CaiSelfDeploy2019}, wearable device~\cite{CaoWear2018}, intelligent robot, autonomous vehicle and unmanned aerial vehicle~\cite{HuMultiDro2018,HuMultiUAV2019}. And an open-source programming framework to support rapid and flexible prototyping and management of human-centered applications is critical~\cite{FortinoProg2013}. In the paper~\cite{EstevaDe2017}, it is pointed out that mobile devices equipped with deep neural network can potentially extend the range of dermatologists outside the outpatient service. It is estimated that by 2023, the number of smart phone users will reach 7.2 billion~\cite{Cerwall2018}. It is possible to provide general diagnosis service with low cost~\cite{ChenUncov2018}.

The automatic recognition of patients' skin conditions may become a good promoter for the cognitive medical monitoring framework. On one side, it can reduce the consumption of resources deployed to the medical industry center, and meanwhile automatically feed back the patients' conditions and service experiences evaluation. On the other side, it properly takes into account the consumption of patients' time and money cost as well as the concerns on privacy. The cloud computing technology deployed on the medical center can solve the problem that the local devices, under the big data environment, have insufficient computing and storage capacities to provide the computation-intensive services~\cite{FortinoBodyCloud2014,WangCrossLayer2019,MalikCloudFog2018}. However, the huge amount of data transmission and communication will cause a consumption of network communication resources, it is still unable to meet the delay-sensitive characteristic of cognitive medical services~\cite{PaceAn2019,ZhangHe2017,MuhammedrEdge2018}. The deployment of edge computing technology on the network edge can solve the pressure caused by the large scale of computation-intensive and rich-media tasks~\cite{ChenCoCaCo2018,WangAdapVid2018,DengUtiMaxi2018}. Also, the technology is beneficial to facilitate secure data management and convenient data trading in mobile health care~\cite{LiEdgeCare2019}. Some existing advanced computation offloading schemes such as~\cite{ZhouReOrient2019} and novel routing mechanisms such as~\cite{ChenDomin2015} can be integrated into the system to achieve highly reliable cooperative computing and communication between local terminals and remote clouds~\cite{HuangBuffer2018}.

The traditional skin disease detection system complete the classification output through characteristics extraction of image data set as the input. The existing researches adopt the deep architecture to automate the learning of characteristics~\cite{Dorjskin2018,RezvantalabDe2018,WuWhat2018}, and the priori knowledge based on pathological skin data set is obtained to improve the accuracy of automatic classification. Esteva et al~\cite{EstevaDe2017} put forward the adoption of Deep Convolutional Neural Networks to classify skin diseases, and demonstrate the achievement of expert-level diagnosis. In the literature~\cite{TajbakhshConvo2016}, it is discussed that the pre-trained deep neural network model has an effect superior to the model trained from the beginning, and the problem of insufficient labeled image data of skin disease can be solved by pre-training the Convolutional Neural Network (CNN) with the images from other medical fields.

Due to the limited intelligence of current system, a one-time testing result be concluded by inputting the collected users' skin image data into the system, but the function of monitoring the changes of skin conditions cannot be realized. Meanwhile, current system is a centralized system with a static and centralized database required an active update by expert, which limits the user mobility and cannot realize convenient and high-efficiency self-checking. In addition, the centralized system is unable to provide sufficient resources to support the individualized database for different population groups. Due to the centralization of the database, it is unable to give a good judgment for paroxysmal diseases.

We consider that the future skin disease monitoring system will meet following characteristics:
\begin{itemize}
  \item Real-time: The user's individual database keeps accumulating and storing. The system analyzes user's skin state based on personal historical data and current data, and monitors skin state changes regularly. The camera on smart terminal capture user's skin image, and skin analysis reports feed back to terminals. Users record their skin state changes based on reports.
  \item Dynamic: The physical location of users often changes dynamically, but mobile devices are relatively static for skin disease detection. Through mobile terminals, users can easily and efficiently collect individual skin images. With the computing power of terminals, fast analysis results are provided to users.
  \item Sharing mode: User's skin images can be locally stored for analysis. Multiple users also send their data to the cloud for sharing. The cloud collects data from different users, and conducts more accurate analysis with its powerful storage and computing capacity.
\end{itemize}

Different from the traditional open-ended input/output system, we intend to build a user-centered close-loop system, and consider connecting the mobile terminal users' personal data collection, the communication between mobile terminals and remote data center, and the real-time update of training model. Based on this, a deep skin disease monitoring system based on edge-to-cloud cognitive medical framework is put forward. Specifically, the contributions of this paper are divided into three points as below:
\begin{enumerate}
\item A medical AI framework based on data width evolution and self-learning is proposed. Under such framework, the process of information interaction between users and terminal devices, and the wide collection of data in the close-loop information flow of user and remote medical data center are considered.
\item A data set filter algorithm based on information entropy is given, so as to lighten the load of no-label data sets in terminals and edge cloud in the meantime of improving the data quality of remote cloud data base and the learning ability of analysis model.
\item A load module specially for analysis algorithms is designed. Under such module, it can be compatible with the application requirements according to the learning model selected. Meanwhile, three learning models are deployed successively in the load module and the training process is completed.
\end{enumerate}

The remained parts of this paper are organized as below. In Section~\ref{sec:architecture}, the AI medical framework is presented, and the entities in and function of the AI medical framework are introduced in detail. In Section~\ref{sec:WidthandDeepth}, based on the framework raised, the data width collection and the self-learning process are elaborated. In Section~\ref{sec:CNNmodel}, the training details of the three deep learning models deployed on cloud are provided, including the data set acquisition, model building and training precision. In Section~\ref{sec:testplatform}, the skin disease recognition prototype system is demonstrated, the specific scene case is given, and the computing and communication delay of the system are analyzed. Finally, the whole paper is summarized and future works are discussed in Section~\ref{sec:conclusion}.

\section{Medical AI Framework}
\label{sec:architecture}
\begin{figure}
\centering
\includegraphics[width=\columnwidth]{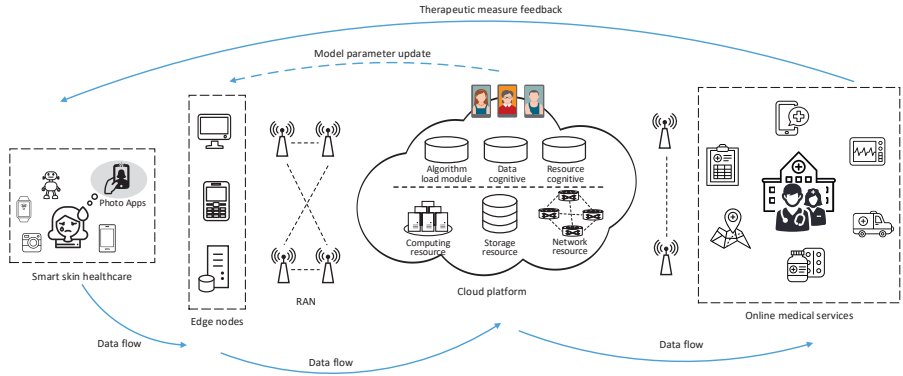}
\caption{The Proposed Medical AI Framework.}
\label{fig001}
\end{figure}

The medical AI framework based on data width evolution and self-learning is shown as Fig.~\ref{fig001}. The framework contains user terminal, edge nodes, radio access network (RAN), cloud platform and remote medical site. Try to imagine an application scenario like this. A user finds abnormal changes in facial skin tissue and has plagued. At this time, the user can, on his/her mobile devices such as mobile phone, easily send the skin images to the edge nodes through taking photos by camera or uploading images from mobile phone gallery. The edge nodes, after data filtering, transmit the skin images to the cloud through the RAN. The cloud provides analysis results on the user's skin conditions based on the deployed learning model, and meanwhile transmits the results to the remote medical site. Upon receipt of the user's data, the specialist physician feeds back the medical measures to the user, and meanwhile archives the medical records to evaluate the changes of the user's skin conditions. The main parts involved in the framework are introduced in details.

\subsection{User terminal} It refers to users' terminal devices, mainly including smart phone, smart bracelet, camera, humanoid robot and other intelligent devices. The terminal device itself contains the data storage module, data sending module, data processing module, and data receiving module. The user, after collecting his/her skin images through the device's shooting App or from the device gallery, firstly makes simple pre-processing by the data processing module and then uploads the skin images to edge node by the data sending module, and the data receiving module will receive the medical feedback data transmitted from the remote cloud or remote site. These devices are characterized by high mobility, low computing resources, and low storage resources. To deal with wireless transmissions in high mobility situations such as vehicular environments, we can employ existing self-organized cooperative transmission scheme like~\cite{TianSeOrga2017} or some advanced routing protocols~\cite{TianMicro2018}.

\subsection{Edge node} It refers to node equipment deployed on the network edge with relatively high computing resources and storage resources, such as the local server. The local server has deployed the learning model, which can carry out skin condition recognition according to users' local data. The edge node transmits the skin image to the remote cloud through RAN, and meanwhile receives the updating parameters of the trained learning model from remote cloud, so as to provide high-efficiency local services.

\subsection{Cloud platform} The cloud platform provides computation-intensive task processing services. The learning ability of model in algorithm load module is improved through receiving the skin image data from the edge nodes. The algorithm load module realizes the adaptation with the application requirements by loading different learning models, and the updated model parameters are transmitted to the edge nodes. The resource cognition module cognizes the network resources, and the data cognition module cognizes the application context and network environment context. The two modules act upon each other to carrying out the network resource management and allocation to meet service requirements of applications.

\subsection{Remote medical site} It refers to the remote medical resources, including doctors, nurses, medical devices and etc. The remote dermatologists receive the user skin conditions analysis results from the cloud, provide online medical services and feed back to user terminals. Users receive the suggested treatment from remote dermatologists, and meanwhile evaluate the service contents.

\section{Data Width Evolution and Self-learning Process}
\label{sec:WidthandDeepth}
\begin{figure}
\centering
\includegraphics[width=\columnwidth]{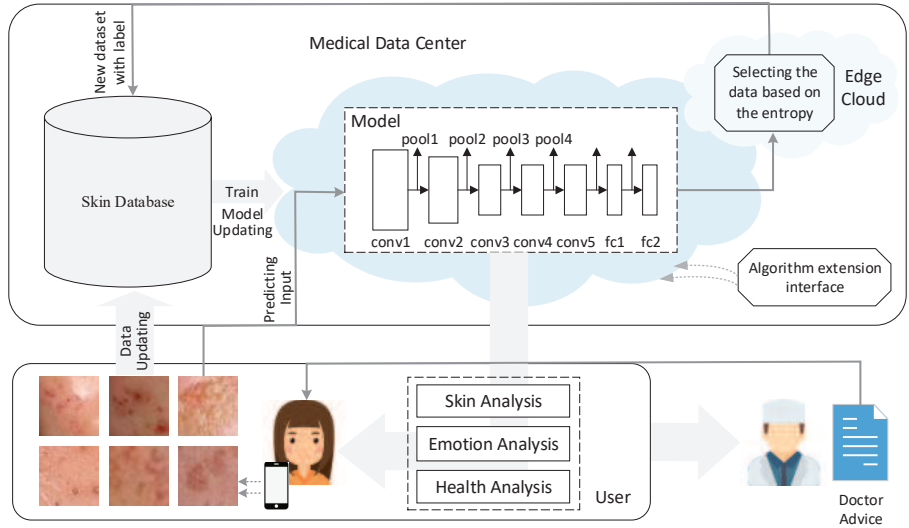}
\caption{The Illustration of Data Width Collection and Self-learning Process.}
\label{fig002}
\end{figure}

Next, we discuss the close-loop data flow in the framework, and give the data width collection and the self-learning process. It is shown as Fig.~\ref{fig002}.

\subsection{Data Width collection}
First, the terminal devices acquire users' skin images and transmit them to the remote medical cloud platform, and the cloud provides skin disease diagnosis service for users by the traditional method based on skin database and deep learning. When the edge nodes receive the continuously accumulated image data from user, the edge cloud pre-processes the local data based on the local cognition, and then send the user data to the remote medical cloud platform. The remote cloud receives the data set from multiple terminals and deep model parameters updated based on global cognition, and then further feeds back the updated parameters to the edge node for a better local cognition. In the whole close-loop process, the skin image data of users, the local cognition data of edge nodes, and the global cognition of the cloud are transmitted and communicated mutually, so as to continuously explore valuable information. In addition, the remote cloud cognizes users' skin conditions according to the deep learning model, and meanwhile judges users' health condition and emotional state and feed back to users. While requesting services, users may also provide the skin condition data, health condition data, emotional state data and surrounding environment information for remote intelligent analysis. The use value of the framework can be expanded horizontally through continuous infusion of information based on user, environment and model into the system.

\subsection{Self-learning Process}
After continuous infusion of information based on user, environment and model into the system, the huge amount of data loaded in the system are unlabeled. It cannot be guaranteed that the unlabeled data set plays a positive role in the model training and global cognition of the cloud platform. Meanwhile, the transmission of huge amount of unlabeled data set will consume the network communication resources and thus reduce the service experience of users~\cite{ChenData2018}. Deploy a data filter algorithm in the edge cloud to filter out the worthless data, and upload the valuable data to the cloud~\cite{ChenLabel2018}. The edge cloud, based on the information entropy, filters and provides the valuable data to the remote cloud. The remote cloud further adjusts and optimizes the model parameters to longitudinally explore more valuable information.

We assume that the labeled data sets are $x^l=[x_1^{l},x_2^{l},\cdots,x_n^{l},\cdots,x_n^{l}],(1\leq n\leq N)$, where $N$ is the number of labeled data sets. The label classes corresponding to the labeled data sets are $y^l=[y_1^{l},y_2^{l},\cdots,y_m^{l},\cdots,y_M^{l}],(1\leq m\leq M)$, respectively, where $M$ is the number of label classes, and for binary classification problem, $M=2$ . Assume that the unlabeled data sets are $x^u=[x_1^{u},x_2^{u},\cdots,x_k^{u},\cdots,x_K^{u}],(1\leq k\leq K)$, where $K$  is the number of unlabeled data sets. We consider making skin color classification on the unlabeled data sets which is denoted as $c^u=[c_1^{u},c_2^{u},\cdots,c_s^{u},\cdots,c_S^{u}],(1\leq s\leq S)$, where $S$ denotes the number of skin color classifications. Then with the conditions of $x_i^{u}$  already labeled as class $c_s$, the probability of being predicted as skin disease class $j$ is $p_i^{j}=p\left(y_{x_i^{u}}=j|c_s\right)$, and it is concluded that the prediction probability of $x_i^{u}$ is $p_{x_i^{u}}=\{p_i^{1},p_i^{2},\cdots,p_i^{M}\}$. On this basis, the prediction probability entropy of unlabeled data is defined as:
\begin{equation}\label{Equation15}
     E\left(p_{x_i^u}\right)= E\left(y_{x_i^{u}}=j|c_s\right)=-\sum_{j=1}^{M}p_i^{j}\log\left(p_i^{j}\right).
\end{equation}

If the entropy value is less than a certain threshold value $E_T$, i.e. $E\left(p_{x_i^u}\right)<E_T$, the unlabeled data is selected. The threshold value $E_T$ is related to the sample size of labeled data, the accuracy rate of model classification, and the quality of service required by users. When the entropy value is relatively small, the newly selected data has a lower prediction uncertainty.

A large amount of unlabeled data collected from users' personal terminal devices is stored in the edge cloud. The filtration of the unlabeled data in the edge nodes decreases the transmission amount of image data with low value, and reduces the communication delay of user service. Meanwhile, owe to the preliminary filtration operation in edge nodes, the cloud utilizes the most valuable data to update the knowledge base, which guarantees the precision of classification. In addition, the data selection according to the skin color labeled by users can form new database classifying by population groups, so as to support the individualized database for different groups.

\section{CNN Model Training and Comparison}
\label{sec:CNNmodel}
A data set used for the classification of human face skin disease is built. The human face images on web pages are crawled by keyword search. The keywords are human face pictures, human face skin disease. The first 20 pages of dynamic web pages are selected for each keyword. Totally 6,144 images are obtained through crawling. The dermatologists from Wuhan Union Hospital are invited to classify all skin images. The labels of images contain 14 classes, including facial acnes, forehead acnes, alar acnes, acne marks, chloasma, pregnant spots, sunburn spots, radiation spots, age spots, dark circles, blackheads, nevus, large pores, wrinkles. Considering the characteristics of diseases, facial acnes, forehead acnes, alar acnes and acne marks are unified as skin acnes. And chloasma, pregnant spots, sunburn spots, radiation spots and age spots are unified as skin spots. Unusable images are removed from the data set. Finally the images are classified as five types of skin diseases. In the classification of skin acnes, skin spots, skin blackheads, dark circles and clean face, the images having skin acnes are taken as the positive sample, and others having skin spots, skin blackheads, dark circles and clean face are taken as the negative samples. In the selection of negative samples, the number of images for each disease type is selected according to the number of images for positive sample, so as to make the ratio of positive samples and negative samples is about 1:1, which avoids the problem of sample imbalance.

\begin{figure}
\centering
\includegraphics[width=\columnwidth]{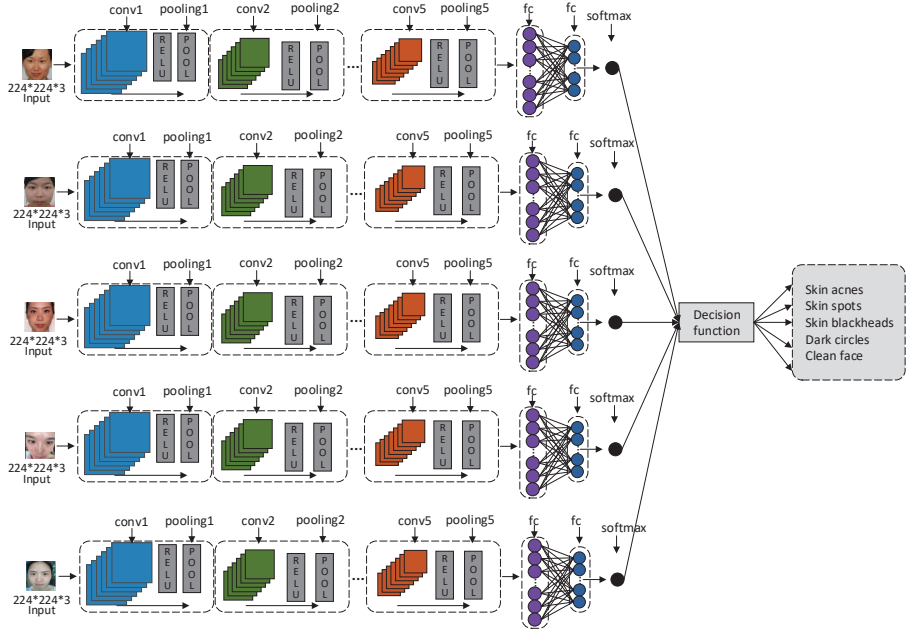}
\caption{The Diagram of CNN Model for Skin Disease Classification.}
\label{fig003}
\end{figure}

The diagram of CNN model for skin disease classification is shown as Fig.~\ref{fig003}. The system uses the Convolutional Neural Network model to extract skin image characteristics. Three learning models, i.e. LeNet-5~\cite{LeCunGra1998}, AlexNet~\cite{KrizhevskyImage2012} and VGG16~\cite{SimonyanVery2014}, are adopted to carry out the training, classification and assessment processes. In the experiment, 85\% of the data set is used as the training set, and the remaining 15\% is used as test set.

Firstly, three deep Convolutional Neural Networks are pre-trained on ImageNet~\cite{Russakovsky2015}. Then a fine tuning is carried out on all layers. The last layer is the softmax layer, which allows to do classification on two diagnosis classes. LeNet-5 is set as 2 convolutional layers, 2 max-pooling layers and 3 fully connected layers. The sizes of all input images are adjusted as 228*228*3, and the input images are normalized. AlexNet is set as 5 convolutional layers, 3 max-pooling layers and 3 fully connected layers. The max-pooling operation is carried after the 1st, 2nd and 5th convolution. The sizes of all input images are adjusted as 227*227*3, and the input images are normalized. The learning rate in the experiment is set as 0.001, and the number of iterations is 150. The batch sizes of the training set and test set are respectively 64 and 5, and the dropout rate is 0.6. VGG16 is set as 13 convolutional layers, 5 max-pooling layers and 3 fully connected layers. The sizes of all input images are adjusted as 227*227*3, and the input images are normalized. The batch size of the training set is 32. The number of iterations is 200. Other parameters are set as the same with LeNet-5 and AlexNet. The accuracy rates of the three learning models on five classes of skin disease are shown as Table.~\ref{tab.notations1}. It can be seen from the statistical data that the AlexNet model has the best overall effect.

\begin{table*}[ht]
\centering
\caption{Skin Disease Classification Accuracy Rate of Three CNN Models.}
\label{tab.notations1}
\scriptsize
\begin{tabular}{|c|c|c|c|c|c|}
\hline
&Skin acnes	&Skin spots	&Skin blackheads	&Dark circles	&Clean face\\
\hline
LeNet-5 &0.63   &0.65   &0.70   &0.58   &0.87\\
\hline
AlexNet	&0.79	&0.80	&0.91	&0.78	&0.95\\
\hline
VGG16	&0.68	&0.75	&0.87   &0.76	&0.90\\
\hline
\end{tabular}
\end{table*}

\section{A demonstration system for skin disease recognition}
\label{sec:testplatform}
\subsection{Prototype Platform}

\begin{figure}
\centering
\includegraphics[width=\columnwidth]{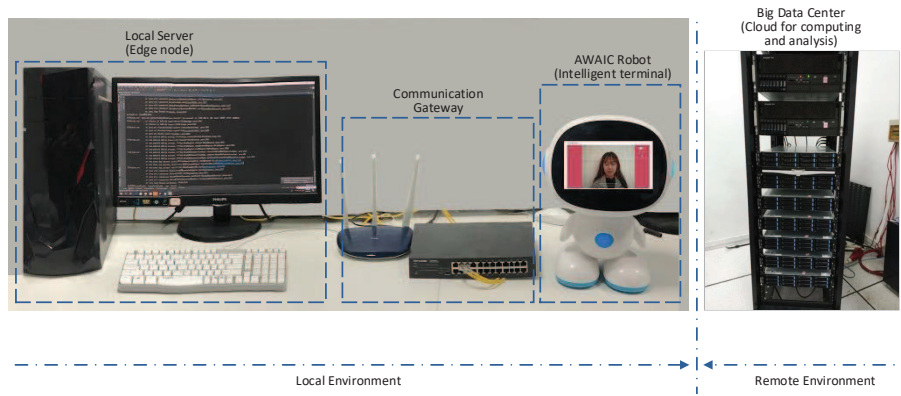}
\caption{The AI-Skin Prototype Platform.}
\label{fig004}
\end{figure}

AI skin disease recognition prototype platform is built, as shown in Fig.~\ref{fig004}. The hardware environment includes AIWAC (Affective Interaction through Wide Learning and Cognitive Computing) robot~\cite{ChenWe2015}, local server, and remote cloud platform. The AIWAC robot is equipped with AI skin App, and captures the user image through the camera above the display screen. The local server is the edge node. The remote cloud is equipped with AMD FX 8-Core processor in 4GHz with 32 GB RAM DDR3. Under current environment, the communication gateway is the communication bridge for the AIWAC robot and local server, the local server and remote cloud, and the remote cloud and AIWAC robot.

The AI skin disease detection processes are as follows: Firstly, the training of skin disease classification models is executed on the remote algorithm server based on our own skin database. After the completion of training, the trained models are stored in the cloud platform, and meanwhile migrated to local server for execution. Then, the skin images of user captured from AWAIC robot are transmitted to edge node. When edge node receives those data, the unlabeled data are selecting based on data set filter algorithm, and then labeled together with the recognition model. The labeled data transmitted to remote cloud for deep training and the updated model parameters are fed back to the edge node. The algorithm extension interface deployed in cloud server can load different algorithm models. To execute the recognition algorithm in edge device, it is required to deploy TensorFlow environment~\cite{tersorflow}.

\subsection{Test Scene}
\begin{figure}
\centering
\includegraphics[width=4in]{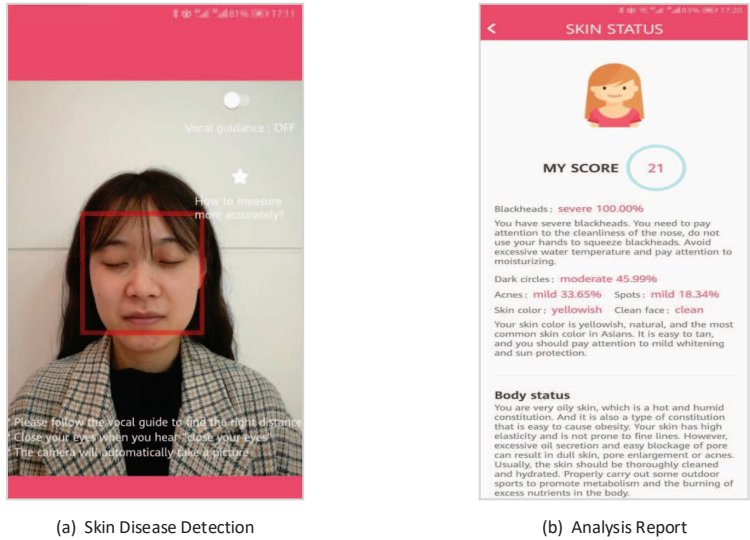}
\caption{The Skin Disease Recognition Demo and Analysis Report.}
\label{fig005}
\end{figure}

An experimental test is carried out on the AI skin prototype platform. The tester captures her own face skin image through the mobile terminal camera, and the OpenCV~\cite{opencv} face detection classifier is utilized to label and segment face area as the input of model. The skin disease recognition algorithm deployed in local server is AlexNet model, which is selected by the optimal result in cloud training. Based on the types of skin diseases given by Table.~\ref{tab.notations2}, the skin conditions of the tester are analyzed for five classes of skin diseases, i.e. (skin acnes, no skin acne), (skin spots, no skin spot), (skin blackheads, no skin blackhead), (dark circles, no dark circle), and (clean face, unclean face). After the completion of skin disease analysis, the skin condition report of the tester is fed back to the mobile terminal. The execution results of real-time analysis of skin conditions are shown as Fig.~\ref{fig005}.

\begin{table*}[ht]
\centering
\caption{Five Types of Skin Diseases Classification.}
\label{tab.notations2}
\scriptsize
\begin{tabular}{|c|c|c|}
\hline
Type    &Class  &Class\\
\hline
1   &Skin acnes	   &No skin acne\\
\hline
2   &Skin spots    &No skin spot\\
\hline
3   &Skin blackheads    &No skin blackhead\\
\hline
4   &Dark circles	&No dark circle\\
\hline
5   &Clean face	&Unclean face\\
\hline
\end{tabular}
\end{table*}

The skin disease recognition report is shown as Fig.~\ref{fig005} (b). First, the system gives an overall score based on the skin condition of the tester and the score is weighted by different skin diseases. Next, the exact analysis result of each disease class is given. It is pointed out that, the problem of blackheads is relatively serious, other skin diseases, including the skin acnes and skin spots are only in mild degree. Moreover, the skin color of the tester is the yellowish skin which is common in Asian people. It can be seen that these results are consistent with the skin state of the tester. In addition, according to the skin analysis, the report indicates that the tester belongs to the damp-heat constitution and the excessive oil secretion leads to the enlarged pores or acnes. The report also proposes measures for improving the skin conditions, such as increasing outdoor sports to promote metabolism.

\subsection{Delay Analysis}
To evaluate the system's reliability and validity, two different models, i.e. LeNet-5 and AlexNet, are compared for the system's computation delay and transmission delay. In the experiment, the communication bandwidth is 2 Mbps.

\begin{figure}
\centering
\includegraphics[width=4in]{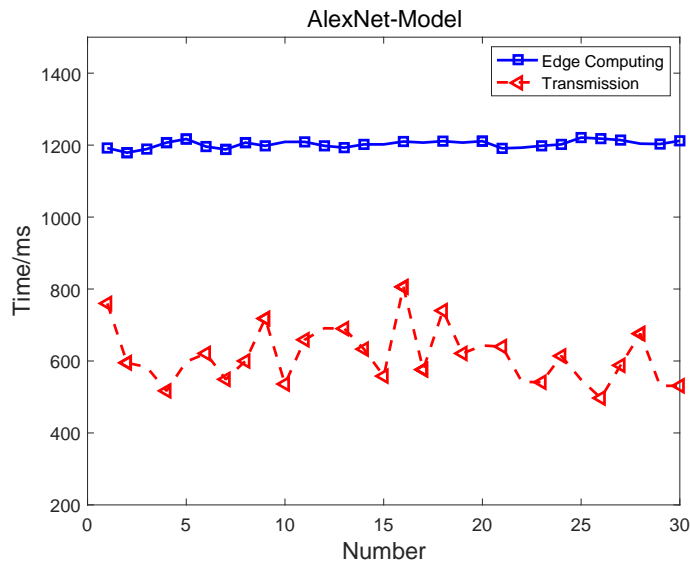}
\caption{Edge Computation Delay and Transmission Delay with AlexNet Model.}
\label{fig006}
\end{figure}

\begin{figure}
\centering
\includegraphics[width=4in]{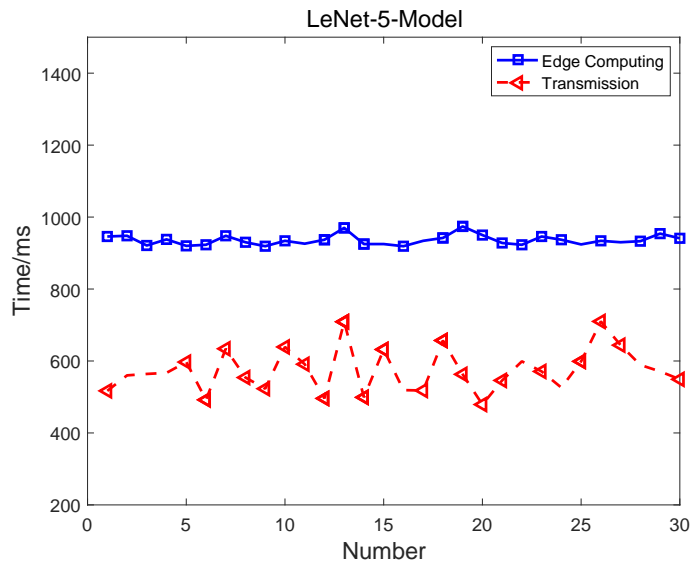}
\caption{Edge Computation Delay and Transmission Delay with LeNet-5 Model.}
\label{fig007}
\end{figure}

We conducted 30 experiments under two models, and the sequence numbers are from 1 to 30. The results of system delay under AlexNet and LeNet-5 models are shown in Fig.~\ref{fig006} and Fig.~\ref{fig007}, respectively. From the figure, we can see that the computation delay of each model changes smoothly with the number of experiments. Without considering the transmission delay of instructions in communication, we can see that the factor determining the total time delay in the system is the computation delay of edge nodes. The average delay of edge computation under AlexNet model is 1.2s, but the end-to-end communication delay (the sum of computation and transmission delay) between terminal device and edge node is still in the order of 1s. The standard deviations of the total communication delay under the two models are 75 ms and 63 ms, respectively, which can show the effectiveness and flexibility of the real-time skin disease recognition system.

\begin{figure}
\centering
\includegraphics[width=4in]{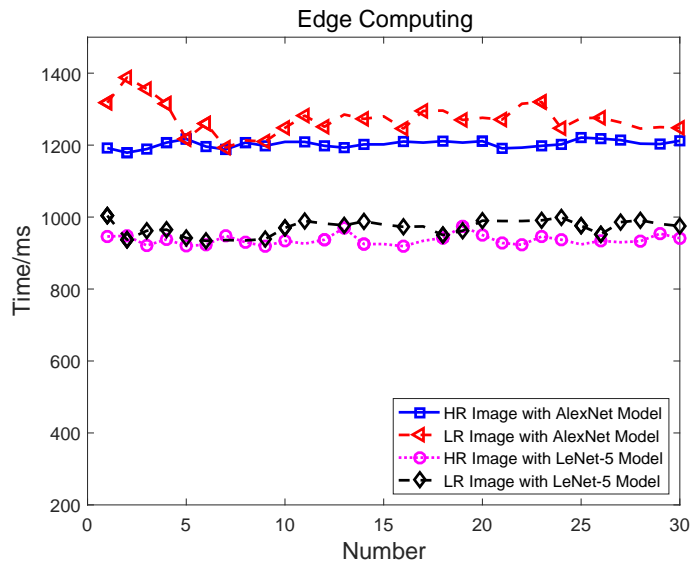}
\caption{Edge Computation Delay of Images with High- and Low- Resolutions.}
\label{fig008}
\end{figure}

Moreover, it is found that the edge computing time of images with high resolution is shorter than that with low resolution, as shown in Fig.~\ref{fig008}. In the experiment, the size of original image is 1233634 bytes, and the size of the compressed image is 4830 bytes. The computing time of high-resolution image is shorter than that of low-resolution image under both AlexNet and LeNet model. The image has a 1.0 probability of detecting skin blackhead disease at both resolutions. Upon testing the images of other skin diseases, it is found that the high- and low-resolution images of blackheads and skin spots have little impact on the classification accuracy.

\section{Conclusion}
\label{sec:conclusion}
In this paper, a real-time, individualized and extensible skin disease recognition system is presented. A medical AI framework based on data width evolution and self-learning is proposed. The close-loop information flow between user and remote medical data center is discussed based on the updating of the data sets such as user's skin images, user's health conditions, environment information, and model parameters in AI skin detection process. In addition, a data set filter algorithm based on information entropy is given. Through the filtration of valuable data sets in the edge node, the data quality of the remote cloud database and the learning ability of models can be further improved. The universality of algorithm extension interface is verified based on the three learning models trained on the cloud, i.e. LeNet-5, AlexNet and VGG16. A skin disease recognition prototype system is built. And skin disease analysis result with the tester's face skin image shot on the mobile terminal camera is conducted. Meanwhile, the edge computation delay and transmission delay of the system is tested, so as to verify the reliability and validity of the system. In our experiment, the end-to-end communication delay between terminal device and edge node is in the order of 1s. We found that the high- and low-resolution images of some skin diseases have little impact on the classification accuracy. There is a tradeoff between transmission delay and classification accuracy. In the future, lower transmission delay can be realized through the deployment of image compression algorithms on terminals~\cite{LiHiera2018}. Moreover, the classification accuracy of skin disease can be further improved by the improvement of learning model~\cite{WangInedge2019}.

\section*{Acknowledgment}

The authors are grateful to the Deanship of Scientific Research at King Saud University for funding this work through the Vice Deanship of Scientific Research Chairs: Chair of Pervasive and Mobile Computing.

\section*{References}


\end{document}